\documentclass[10pt,twocolumn,letterpaper]{article}
\PassOptionsToPackage{bookmarks=false}{hyperref}
\DeclareUnicodeCharacter{2212}{-}

\usepackage{cvpr}      
\usepackage{multirow}
\usepackage[table,xcdraw]{xcolor}
\usepackage{flushend}

\definecolor{cvprblue}{rgb}{0.21,0.49,0.74}
\usepackage[draft]{hyperref}

\def\paperID{18046} 
\def\confName{CVPR}
\def\confYear{2026}

\title{VEMamba: Efficient Isotropic Reconstruction of Volume Electron Microscopy with Axial-Lateral Consistent Mamba}

\author{
Longmi Gao \quad Pan Gao\thanks{Corresponding author.}\\
College of Artificial Intelligence, Nanjing University of Aeronautics and Astronautics\\
Nanjing 211106, China\\
{\tt\small gaolongmi@nuaa.edu.cn, Pan.Gao@nuaa.edu.cn}
}

\begin{document}
\maketitle
\begin{abstract}
Volume Electron Microscopy (VEM) is crucial for 3D tissue imaging but often produces anisotropic data with poor axial resolution, hindering visualization and downstream analysis. Existing methods for isotropic reconstruction often suffer from neglecting abundant axial information and employing simple downsampling to simulate anisotropic data. To address these limitations, we propose VEMamba, an efficient framework for isotropic reconstruction. The core of VEMamba is a novel \textbf{3D Dependency Reordering} paradigm, implemented via two key components: an Axial-Lateral Chunking Selective Scan Module (ALCSSM), which intelligently re-maps complex 3D spatial dependencies (both axial and lateral) into optimized 1D sequences for efficient Mamba-based modeling, explicitly enforcing axial-lateral consistency; and a Dynamic Weights Aggregation Module (DWAM) to adaptively aggregate these reordered sequence outputs for enhanced representational power. Furthermore, we introduce a realistic degradation simulation and then leverage Momentum Contrast (MoCo) to integrate this degradation-aware knowledge into the network for superior reconstruction. Extensive experiments on both simulated and real-world anisotropic VEM datasets demonstrate that VEMamba achieves highly competitive performance across various metrics while maintaining a lower computational footprint. The source code is available on GitHub: \url{https://github.com/I2-Multimedia-Lab/VEMamba}
\end{abstract}    
\section{Introduction}

\begin{figure}[ht]
    \centering
\includegraphics[width=\columnwidth,height=0.35\textheight,keepaspectratio]{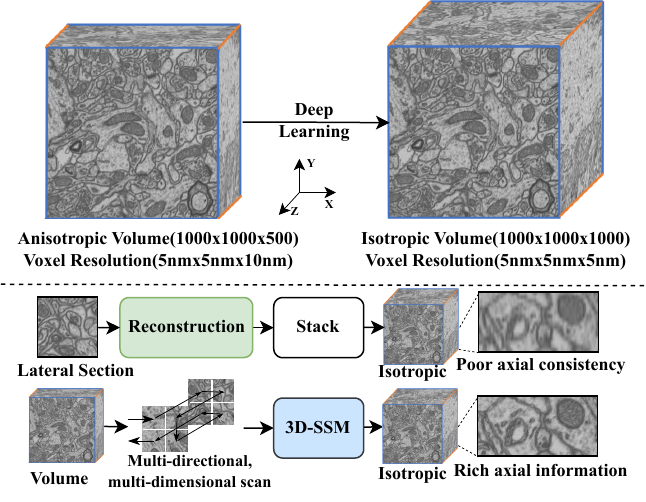}
    \caption{At the top is the Deep Learning for Isotropic Reconstruction. Compared with isotropic techniques, anisotropic techniques produce fewer axial pixels when imaging the same tissue. Deep learning is then used to compensate for the missing axial pixels. At the bottom is the Comparison between the common method and VEMamba. The common method takes lateral sections as input and reconstructs a volume by stacking, leading to poor axial consistency. In contrast, VEMamba takes volumetric input and performs multi-directional, multi-dimensional scanning, yielding rich axial information.
    }
    \label{fig:1}
\end{figure}

Volume Electron Microscopy is a pivotal technology in modern life sciences~\cite{vem7,vem8}, medicine~\cite{vem9,vem10,vem11}, and clinical diagnosis~\cite{vem12,vem13,vem14}, enabling the direct 3D visualization of cellular~\cite{vem1,vem3,vem4} and tissue~\cite{vem5,vem6} ultrastructures at nanometer resolution. Isotropic high-resolution data are crucial for the accurate analysis and comprehensive understanding of complex biological specimens. However, most VEM imaging techniques, such as serial section Transmission Electron Microscopy (ssTEM), inherently produce anisotropic volumes~\cite{ssTEM}. These volumes suffer from lower axial ($z$) resolution compared to the lateral ($x,y$) resolution, a limitation imposed by the physical constraints of serial sectioning, as shown in Figure~\ref{fig:1}. While certain advanced techniques such as focused ion beam scanning electron microscopy (FIB-SEM) can acquire isotropic data, they suffer from prohibitively slow imaging speeds and high costs, making them impractical for widespread adoption~\cite{FIBSEM}. Therefore, developing efficient algorithms for isotropic reconstruction from anisotropic data is essential for advancing biological research.

Early supervised method~\cite{FSRCNN} showed potential, but the difficulty in obtaining isotropic data led to a shift towards self-supervised frameworks. The standard self-supervised paradigm involves training on the high-resolution lateral sections and validating the reconstruction performance on the axial dimension. However, existing self-supervised frameworks, including GAN-based~\cite{SRGAN,cycleGAN} and Diffusion-based~\cite{DiffuseEM,DiffuseIR}, still encounter fundamental limitations. First, the vast majority rely on 2D model architectures. This architectural choice prevents them from effectively modeling the 3D spatial dependency inherent in volumetric data, often leading to artifacts and a lack of coherence between adjacent slices, as shown in Figure~\ref{fig:1}. While some 3D approaches using Transformers~\cite{attention,isovem} have been explored, they often incur prohibitive computational costs and memory footprints, making them intractable for high-resolution volumetric data. Second, the common simulation practice of simple downsampling fails to capture the complex degradations present in real-world acquisitions. This discrepancy leads to suboptimal performance when models are deployed on real anisotropic data.

The recent advent of State Space Models~\cite{ssm1,ssm2,ssm3}, particularly the Mamba~\cite{mamba} architecture, presents a new opportunity to overcome these challenges. Mamba is distinguished by its ability to model global dependencies, while maintaining linear complexity and high memory efficiency~\cite{mamba1,mamba2}. This unique combination makes it exceptionally well-suited for modeling large-scale 3D volume. 
\emph{To the best of our knowledge, we are the first to apply the Mamba architecture for VEM isotropic reconstruction.}

Building on these advantages, we introduce VEMamba, a novel, \emph{efficient} yet \emph{effective} framework for VEM isotropic reconstruction. The core of our VEMamba architecture is composed of two  components: the Axial-Lateral Chunking Selective Scan Module (ALCSSM) and the Dynamic Weights Aggregation Module (DWAM). ALCSSM is the core method for achieving 3D dependency reordering. It reorders and maps the physically separated axial (inter-slice) and lateral (intra-slice) dependencies in a 3D volume into 1D sequences that Mamba can process. By performing orthogonal axial–lateral scans, the SSM operating on these 1D sequences is forced to model both inter-slice and intra-slice information flow simultaneously, thereby explicitly establishing axial–lateral consistency. This process is further enhanced by the DWAM, which adaptively computes dynamic weights for the feature sequences generated by the different scans, enabling a more powerful and context-aware aggregation of the 3D representation. Furthermore, we introduce a  degradation model incorporating multiple types of degradation including blur, downsampling, and noise to more accurately simulate real anisotropic data characteristics and employ MoCo~\cite{moco} for self-supervised extraction of degradation information, enabling better reconstruction of lateral sections. Our contributions can be summarized as follows: 
\begin{itemize}
\item We propose the Axial-Lateral Chunking Selective Scan Module and Dynamic Weights Aggregation Module to implement an efficient 3D dependency reordering paradigm, ensuring robust axial-lateral consistency.
\item We introduce a degradation learning framework that incorporates multiple types of degradation to accurately simulate real anisotropic data, coupled with MoCo-based self-supervised degradation extraction for enhanced lateral section reconstruction.
\item Experiments demonstrate highly competitive performance on both synthetic and real VEM datasets, achieving superior reconstruction quality in the majority of evaluated settings while maintaining computational efficiency.
\end{itemize}

\section{Related Work}
\begin{figure*}[t]
    \centering
    \includegraphics[width=\textwidth]{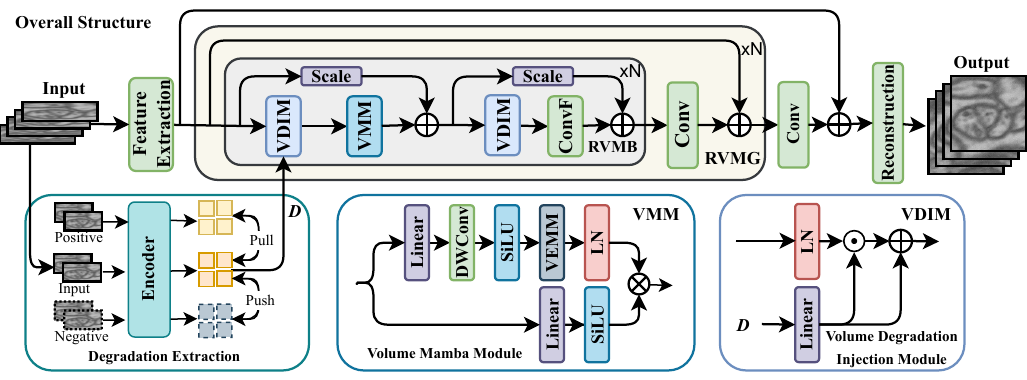} 
    \caption{Visualization of VEMamba. At the top is the overall structure of VEMamba, which is primarily composed of the Residual Volume Mamba Group (RVMG), and the RVMG consists mainly of the Residual Volume Mamba Block (RVMB). On the bottom left is the Degradation Extraction. On the bottom middle is the Volume Mamba Module (VMM). On the bottom right is the Volume Degradation Injection Module (VDIM).}
    \label{fig:main_mamba}
\end{figure*}
\textbf{Isotropic Reconstruction for VEM.} Classical interpolation algorithms, such as bicubic or spline interpolation, offer a fast, training-free solution but often yield suboptimal and blurry results, failing to recover fine structural details.

To alleviate this issue, deep learning-based methods have become prevalent. Generative Adversarial Networks (GANs) were widely explored, leveraging the high-resolution lateral sections as an unpaired supervisory signal to restore the low-resolution axial sections~\cite{cycleGAN,SRGAN}. While promising, these GAN-based approaches are often plagued by training instability and can introduce undesirable artifacts. More recent works have explored advanced network architectures. Transformer-based models, for instance, have shown great potential. A2I-3DEM~\cite{A2I-3DEM} integrates a Vision Transformer (ViT) with a U-Net to effectively capture multi-scale local and global image dependencies. Similarly, IsoVEM~\cite{isovem} employs a combination of attention mechanisms within a U-Net architecture to efficiently model inter-section relationships. Inspired by advancements in video processing, some methods~\cite{z-upscaling_2024,wang_temporal_2021} reframe the problem as video frame interpolation, though they are often limited in their reconstruction factor. Recently, diffusion models~\cite{DiffuseEM,DiffuseIR,EMDiffuse} have emerged as a powerful alternative, demonstrating superior generation quality and training stability compared to GANs. These methods typically train a model on 2D lateral slices and apply it to restore axial slices during the iterative sampling process.

Despite this progress, current methods are still hampered by the two primary challenges discussed in the Introduction: a prevalent reliance on 2D architectures that ignore 3D continuity, and simplistic degradation simulations that create a domain gap. Furthermore, each architecture class presents its own practical challenges: GANs suffer from unstable training, Transformers incur a prohibitive memory footprint and computational cost for large volumes, and diffusion models are notoriously slow and resource-intensive due to their iterative sampling process. Our work is designed to directly address these gaps by introducing a 3D-native, efficient Mamba-based architecture and a more realistic self-supervised framework built upon degradation modeling.

\section{Method}

\subsection{Motivation}
Our method addresses two primary challenges in VEM isotropic reconstruction.

\textbf{Axial-Lateral Consistency.} The key challenge is the severe anisotropy. Existing 2D methods cannot capture long-range dependencies along the axial dimension. By contrast, full 3D methods model axial dependencies but incur prohibitive computational and memory costs. This motivated us to design the VEMamba Module (VEMM), which performs 3D spatial context modeling with linear computational complexity and explicitly enforces axial-lateral consistency, thereby resolving the fundamental trade-off between 2D and 3D approaches. At its core, the ALCSSM creates a seamless flow of information across all spatial dimensions, and the DWAM adaptively fuses these multi-directional representations.

\textbf{Realistic Degradation Modeling.} The second challenge is the simulation gap. Common strategies using simple downsampling fail to capture the complex, real-world degradations inherent in VEM acquisition. This discrepancy limits model robustness. To address this, we introduce a more realistic degradation framework and leverage Momentum Contrast. This allows the network to learn a degradation-aware representation in a self-supervised manner, which is then injected via the VDIM to enhance robustness and achieve a more faithful reconstruction.

\begin{figure*}[t]
    \centering
    \includegraphics[width=\textwidth]{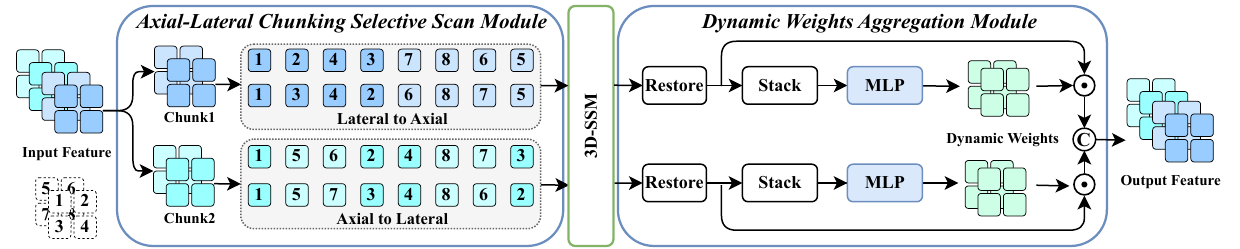} 
    \caption{Visualization of the VEMamba Module (VEMM). It includes our proposed Axial-Lateral Chunking Selective Scan Module (ALCSSM), Dynamic Weights Aggregation Module (DWAM) and SSM.}
    \label{fig:mamba_block}
\end{figure*}

\subsection{Overall Architecture}
Our model is composed of four main stages: a shallow feature extraction stage, a degradation extraction stage, a deep feature extraction stage, and a final reconstruction stage. The overall pipeline is illustrated in Figure~\ref{fig:main_mamba}. Specifically, the input is a single-channel anisotropic subvolume $X \in \mathbb{R}^{F \times h \times W}$, where $F$, $h$, and $W$ represent the number of lateral sections, the height, and the width. Our goal is to reconstruct an isotropic volume $Y \in \mathbb{R}^{F \times H \times W}$, where $H = s \times h$ and $s$ is the axial upscaling factor. Initially, the input volume $X$ is processed by the shallow feature extraction stage to obtain shallow features $F_S \in \mathbb{R}^{F \times C \times h \times W}$, where $C$ denotes the channel dimension of the hidden features. In parallel, the degradation extraction stage uses momentum contrast to obtain a degradation representation $D \in \mathbb{R}^{L}$. The shallow features $F_S$ and the degradation representation $D$ are then fed into the deep feature extraction stage to produce deep features $F_D \in \mathbb{R}^{F \times C \times h \times W}$. This stage is constructed by stacking several Residual Volume Mamba Groups (RVMGs), each containing multiple Residual Volume Mamba Blocks (RVMBs). A RVMB integrates three key components: a Volume Mamba Module (VMM) for capturing global context, a ConvFFN~\cite{convffn} (ConvF) for local details, and a Volume Degradation Injection Module (VDIM) to inject the degradation representation $D$. Finally, the reconstruction stage fuses the feature maps via addition $(F_S + F_D)$ and upscales the result using a pixel-shuffle layer followed by a convolution to produce the final output $Y$.

We optimize the network (except the degradation extraction stage) with a hybrid loss function combining L1 and SSIM losses:
\begin{equation}
\mathcal{L}_{\text{total}} = \mathcal{L}_{1}(Y, \hat{Y}) +  \mathcal{L}_{\text{SSIM}}(Y, \hat{Y})
\end{equation}
where $\hat{Y}$ is the ground-truth high-resolution volume, and $Y$ is the reconstructed volume.

\subsection{VEMamba Module}
The core of VEMamba lies in the VEMamba Module (VEMM). As illustrated in Figure~\ref{fig:mamba_block}, the VEMM operates in three sequential stages: (1) it begins by partitioning the 3D feature map and employing the ALCSSM to transform these partitions into one-dimensional sequences; (2) these sequences are then processed by a SSM to capture long-range dependencies and establish a global receptive field; (3) finally, the resulting sequences are dynamically integrated using our proposed DWAM, ensuring an adaptive and context-aware fusion of multi-directional information.

\textbf{ALCSSM:} The ALCSSM is designed to systematically deconstruct and sequentialize the 3D feature maps in a manner that preserves the crucial interplay between axial and lateral information. This module guarantees a consistent and comprehensive representation of the volumetric data for the subsequent SSM processing. Inspired by the computational efficiency demonstrated in MobileMamba~\cite{mobilemamba}, we initially partition the 3D feature tensor into two distinct chunks along the channel dimension. This strategy not only reduces the GPU memory footprint but also enhances parallel processing capabilities. Subsequently, we introduce a multi-directional, multi-dimensional scanning strategy for each chunk, motivated by the continuous information flow concept in SCST~\cite{SCST}. As depicted in Figure~\ref{fig:mamba_block}, we employ a series of continuous 3D scanning paths that traverse the volume from ``Axial-to-Lateral" and ``Lateral-to-Axial." It is crucial to note that this encompasses not only the two primary directions shown but also their reverse counterparts. This comprehensive set of eight scanning trajectories (four for each chunk) ensures a fluid and uninterrupted flow of information across all spatial dimensions. This synergistic combination of channel partitioning and multi-directional scanning allows ALCSSM to achieve exceptional performance with high computational efficiency, rendering it well suited for the demands of VEM isotropic reconstruction.

\begin{table*}[t]
\centering
\caption{Quantitative evaluation of VEMamba. Comparison of VEMamba and Baseline (interpolation), IsoVEM, and EMDiffuse on the EPFL and CREMI datasets, using PSNR, SSIM, and LPIPS. The table also includes a comparison of computational cost and parameter count. The best results are highlighted in {\color[HTML]{FF0000}red}, and the second-best results are in {\color[HTML]{4472C4}blue}.}
\label{table:1}
\begin{tabular}{@{}ccccccccc@{}} %
\toprule
                                &                             & \multicolumn{3}{c}{EPFL Dataset}                                                                               & \multicolumn{3}{c}{CREMI Dataset}                                                                              &                                             \\
\multirow{-2}{*}{Method}        & \multirow{-2}{*}{Metrics}   & $\times 4$                                     & $\times 8$                                     & $\times 10$                                  & $\times 4 $                                    & $\times 8$                                     & $\times 10$                                    & \multirow{-2}{*}{\begin{tabular}[c]{@{}c@{}}Flops(T) \\ Params(M)\end{tabular}} \\ \midrule %
                                & PSNR                        & 28.407                                 & 25.558                                 & 24.583                                 & 34.219                                 & 28.129                                 & 26.554                                 &                                             \\
                                & SSIM                        & 0.7478                                 & 0.6323                                 & 0.5891                                 & 0.9133                                 & 0.7861                                 & 0.7379                                 &                                             \\
\multirow{-3}{*}{Baseline}      & LPIPS                       & 0.2709                                 & 0.3955                                 & 0.4398                                 & 0.2811                                 & 0.4445                                 & 0.5031                                 & \multirow{-3}{*}{\begin{tabular}[c]{@{}c@{}}-\\ -\end{tabular}} \\ %
                                & PSNR                        & {\color[HTML]{4472C4} 29.234}          & {\color[HTML]{4472C4} 27.119}          & {\color[HTML]{4472C4} 26.138}          & {\color[HTML]{FF0000} 37.807}          & {\color[HTML]{4472C4} 31.449}          & {\color[HTML]{4472C4} 29.662}          &                                             \\
                                & SSIM                        & {\color[HTML]{4472C4} 0.7703}          & {\color[HTML]{FF0000} 0.6839}          & {\color[HTML]{FF0000} 0.6504}          & {\color[HTML]{4472C4} 0.9485}          & {\color[HTML]{4472C4} 0.8647}          & {\color[HTML]{4472C4} 0.8296}          &                                             \\
\multirow{-3}{*}{IsoVEM}        & LPIPS                       & {\color[HTML]{4472C4} 0.2196}          & {\color[HTML]{4472C4} 0.3192}          & {\color[HTML]{4472C4} 0.3546}          & 0.2089                                 & 0.3208                                 & 0.3618                                 & \multirow{-3}{*}{\begin{tabular}[c]{@{}c@{}}{\color[HTML]{4472C4} 0.61} \\ {\color[HTML]{4472C4} 1.40}\end{tabular}} \\ %
                                & PSNR                        & 27.522                                 & 24.947                                 & 24.407                                 & 36.305                                 & 29.027                                 & 27.131                                 &                                             \\
                                & SSIM                        & 0.6679                                 & 0.5733                                 & 0.5508                                 & 0.9326                                 & 0.7955                                 & 0.7214                                 &                                             \\
\multirow{-3}{*}{EMDiffuse}     & LPIPS                       & 0.3716                                 & 0.4938                                 & 0.4596                                 & {\color[HTML]{FF0000} 0.1362}          & {\color[HTML]{FF0000} 0.2418}          & {\color[HTML]{FF0000} 0.3004}          & \multirow{-3}{*}{\begin{tabular}[c]{@{}c@{}}22.51 \\ 15.16\end{tabular}} \\ \midrule %
                                & PSNR                        & {\color[HTML]{FF0000} 29.442}          & {\color[HTML]{FF0000} 27.418}          & {\color[HTML]{FF0000} 26.473}          & {\color[HTML]{4472C4} 37.785}          & {\color[HTML]{FF0000} 31.556}          & {\color[HTML]{FF0000} 29.796}          &                                             \\
                                & SSIM                        & {\color[HTML]{FF0000} 0.7707}          & {\color[HTML]{4472C4} 0.6816}          & {\color[HTML]{4472C4} 0.6487}          & {\color[HTML]{FF0000} 0.9869}          & {\color[HTML]{FF0000} 0.9498}          & {\color[HTML]{FF0000} 0.9278}          &                                             \\
\multirow{-3}{*}{\textbf{Ours}} & LPIPS                       & {\color[HTML]{FF0000} 0.2176}          & {\color[HTML]{FF0000} 0.3006}          & {\color[HTML]{FF0000} 0.3388}          & {\color[HTML]{4472C4} 0.2021}          & {\color[HTML]{4472C4} 0.3129}          & {\color[HTML]{4472C4} 0.3523}          & \multirow{-3}{*}{\begin{tabular}[c]{@{}c@{}}{\color[HTML]{FF0000} 0.28} \\ {\color[HTML]{FF0000} 0.94}\end{tabular}} \\ \bottomrule %
\end{tabular}
\end{table*}

\textbf{DWAM:} Following the global modeling by the SSM, the DWAM is designed to adaptively fuse the sequences generated from the diverse scanning directions. This module is critical for synthesizing the complementary information captured by each scan, thereby enabling a more robust and holistic representation of the 3D volume. As illustrated in Figure~\ref{fig:mamba_block}, let the set of eight 1D sequences processed by the SSM be denoted as $\{x_{i}\}_{i=1}^{8}$. The fusion process commences by first restoring these sequences to their original 3D spatial arrangement, $F_{1}$ and $F_{2}$, corresponding to the two initial chunks:
\begin{equation}
\begin{aligned}
F_{1} &= \mathrm{Restore}(\{x_{i}\}_{i=1}^{4})\\
F_{2} &= \mathrm{Restore}(\{x_{i}\}_{i=5}^{8})
\end{aligned}
\end{equation}

These aggregated feature tensors are stacked to form two distinct 3D feature tensors and then passed through an MLP to generate context-dependent weights, $W_{1}$ and $W_{2}$. These weights dynamically modulate the importance of each feature map based on its contribution to the final reconstruction:
\begin{equation}
\begin{aligned}
W_{1} &= \text{MLP}(\mathrm{Stack}(F_{1})) \\
W_{2} &= \text{MLP}(\mathrm{Stack}(F_{2}))
\end{aligned}
\end{equation}

Finally, the output of the VEMM is obtained by performing an element-wise multiplication of the features with their corresponding adaptive weights and concatenating the results:
\begin{equation}
\mathrm{Out} = \mathrm{Concat}(W_{1} \odot F_{1},\, W_{2} \odot F_{2})
\end{equation}

This adaptive weighting mechanism empowers the model to dynamically emphasize features from the most informative scanning paths based on the input context.

\subsection{Momentum Contrastive Learning}
To enhance the model's robustness and ensure a more faithful reconstruction of lateral information, we introduce an unsupervised strategy to learn and inject degradation representations into our network.

\textbf{Degradation Representation Learning:} Our objective is to learn a latent representation of the degradation information in an unsupervised manner. Specifically, we create a positive pair by sampling two subvolumes along its axial dimension. Subvolumes from different parent volumes form negative pairs. This process trains an encoder to capture implicit degradation features, optimized by the InfoNCE~\cite{infonce} loss. Given a batch of $N$ volumes, each representing a unique degradation profile. From each volume $i$, we extract two subvolumes and encode them into a query $q_{i}$ and a positive key $k_{i+}$. The remaining $N-1$ keys from other volumes serve as negative samples. The training objective is thus as follows:
\begin{equation}
\mathcal{L}_{\text{D}} = - \sum_{i=1}^{N} \log 
\frac{\exp ( q_i \cdot k_{i+} / \tau )}
{\exp ( q_i \cdot k_{i+} / \tau ) + \sum_{\substack{j=1\\ j \neq i} }^{N} \exp ( q_i \cdot k_{j-} / \tau )}
\end{equation}
where $\tau$ is a temperature hyper-parameter.

\textbf{VDIM:} Upon learning the degradation representation $D$, we integrate this information into our main reconstruction network. The VDIM adaptively modulates the network's features by applying channel-wise affine transformations:
\begin{equation}
F' = \text{Linear}(D) \odot \text{Norm}(F) + \text{Linear}(D)
\end{equation}
Here, $F$ and $F'$ represent the input and output feature, respectively.

By modeling and injecting this degradation-aware prior knowledge into the reconstruction network, our framework can perform a more targeted and precise restoration.

\begin{figure*}[t]
    \centering
    \includegraphics[width=\textwidth]{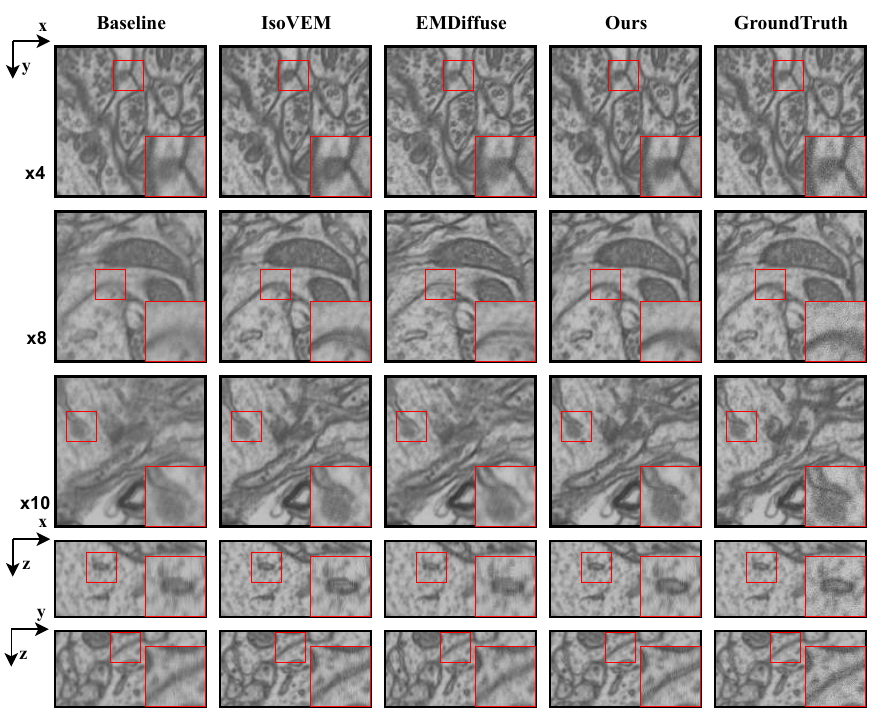} 
    \caption{Qualitative comparison of baseline (Interpolation), IsoVEM, EMDiffuse, VEMamba, and Ground Truth on the EPFL dataset at scale factors of ×4, ×8, and ×10 in the lateral (xy) and axial (xz and yz) section.}
    \label{fig:main1}
\end{figure*}

\begin{figure*}[t]
    \centering
    \includegraphics[width=\textwidth]{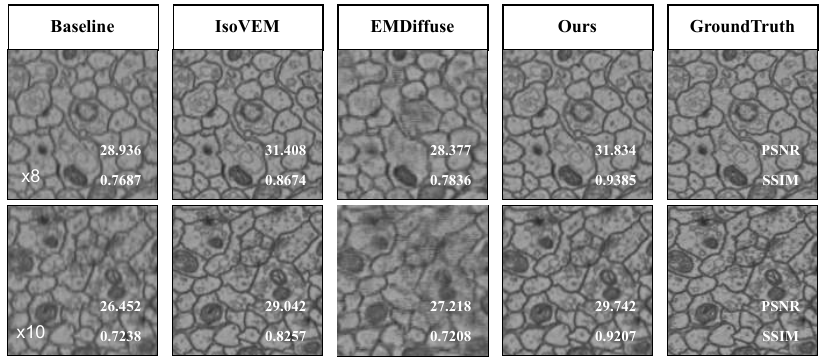} 
    \caption{Visual comparison of baseline (Interpolation), IsoVEM, EMDiffuse, VEMamba, and Ground Truth on the CREMI dataset at scale
factors of ×8 and ×10 in the lateral (xy) section.}
    \label{fig:main2}
\end{figure*}
\section{Experiments}

\subsection{Experimental settings}

\textbf{Implementation Details.} Our proposed VEMamba was implemented in PyTorch and trained on two NVIDIA RTX 4090 GPUs. We employed the Adam~\cite{adam} optimizer with a learning rate of $5\times10^{-5}$, which was adjusted by a cosine annealing scheduler following a 10 epoch warmup period. The model typically converged after 200 epochs. For training, we used a batch size of 2. The input subvolume dimensions were set to (32,128,128) for a scale factor of $\times 4$, (16,128,128) for $\times 8$, and (16,160,160) for $\times 10$. Our baseline model consists of 4 stacked RVMG blocks, with each RVMG block containing 4 RVMBs. The specific parameters for the degradation process (including blur, downsampling, and noise) follow the settings used in DiffuseEM~\cite{DiffuseEM}.

\textbf{Datasets.} We validated our method on both isotropic and anisotropic datasets. For the isotropic dataset, we used the EPFL dataset~\cite{epfl}, which contains a $5 \times 5 \times 5~\mu\mathrm{m}^{3}$ volume of neural tissue from the hippocampal CA1 region acquired via FIB-SEM at a high resolution of $5 \times 5 \times 5~\mathrm{nm}^{3}$/voxel. To evaluate isotropic reconstruction, we synthetically generated anisotropic data by applying $\times 4$, $\times 8$, and $\times 10$ degradations along the z-axis. The CREMI dataset~\cite{cremi} consists of serial section TEM images of an adult Drosophila melanogaster brain. This dataset comprises three samples (A, B, and C), each with 193 sections at a resolution of $4 \times 4 \times 40~\mathrm{nm}^{3}$/voxel. Due to the sparse axial sampling, we conducted training and evaluation on the lateral planes for this dataset.


\subsection{Experimental Results}
To ensure a fair and rigorous comparison, we retrained IsoVEM and EMDiffuse on our experimental datasets, as their original publications either did not report metrics on these specific datasets or used different datasets. VEMamba demonstrates robust and highly competitive performance across both the EPFL and CREMI datasets.

\textbf{Quantitative Comparison. }The comprehensive quantitative results are presented in Table~\ref{table:1}. In terms of PSNR, VEMamba achieves the top result in five out of six experimental settings. In these cases, it surpasses the second-best method by a significant margin of approximately 0.2-0.3 dB. In the single instance where our model ranked second, the performance gap was merely 0.022 dB, underscoring its consistent and robust performance. Similarly, for the SSIM, VEMamba achieves the best results in four of the six experiments, again outperforming competitors by a substantial margin. For the two settings where it was the runner-up, the differences were minimal, at just 0.0023 and 0.0017, respectively. This highlights our model's superior capability in preserving structural details during reconstruction. Furthermore, VEMamba shows highly competitive results on the LPIPS metric. Although it achieves the top score in three settings, it is worth noting that the LPIPS metric is pre-trained on RGB natural images. Its direct application to grayscale electron microscopy images may introduce a slight domain bias. Nevertheless, the strong performance indicates that our reconstructions are perceptually closer to the ground truth.

\textbf{Qualitative Comparison. }Visual comparisons provide further evidence of VEMamba's superiority, as illustrated in Figure~\ref{fig:main1} for the lateral (xy) plane and the axial (xz and yz) planes. In the xy-plane reconstructions at a 4x scale factor, IsoVEM tends to hallucinate non-existent boundaries and textural details when compared to the GT. On the other hand, EMDiffuse struggles with reconstruction fidelity, failing to restore fine tissue structures accurately. In stark contrast, our model's output is most faithful to the GT. This trend persists at higher magnification factors. At 8x, IsoVEM generates spurious membrane structures, while EMDiffuse fails to reconstruct them at all. VEMamba delivers the most accurate result. At the 10x scale, both IsoVEM and EMDiffuse are unable to produce complete and coherent tissue boundaries, whereas our method reconstructs them with remarkable clarity and integrity. The qualitative results in the xz and yz planes mirror these findings. Notably, in the xz-plane, IsoVEM produces prominent artifacts, which are particularly detrimental to achieving true isotropic reconstruction. The visualization results on the CREMI dataset are similar to those described above, as illustrated in Figure~\ref{fig:main2}. Across all visual comparisons, VEMamba consistently yields the highest-quality reconstructions, free from the artifacts and structural inconsistencies that plague other methods. These qualitative results further validate the efficacy of our designed modules in implementing the proposed 3D dependency reordering.

\textbf{Computational Cost. }As detailed in Table~\ref{table:1}, VEMamba is not only effective but also highly efficient. Our model features the lowest parameter count and computational load among all compared deep learning models. This combination of superior quantitative and qualitative performance with minimal resource requirements demonstrates the efficacy and efficiency of our design. The lightweight nature of VEMamba facilitates easier deployment in real-world scenarios, offering a practical solution to enhance the imaging capabilities of volume electron microscopes and thereby advancing biomedical research.

\subsection{Downstream Task}
To assess the practical utility of our reconstruction, we evaluate its performance on a critical downstream task: mitochondria segmentation. High-quality isotropic reconstruction should not only exhibit high visual quality but also improve the accuracy of subsequent quantitative analyses, such as organelle segmentation, which is vital for biological studies. We employ a U-Net~\cite{unet} architecture for segmentation, utilizing the Intersection over Union (IoU) as the primary evaluation metric. The network is trained for 200 epochs with the Adam optimizer, a learning rate of \(1 \times 10^{-4}\), and the BCEWithLogitsLoss function.

As presented in Table~\ref{table:seg}, our proposed VEMamba consistently achieves state-of-the-art performance across all scale factors ($\times$4, $\times$8, and $\times$10). Notably, the performance gap between our reconstruction and the isotropic ground truth is remarkably narrow, with an IoU difference of merely 0.002. 

\begin{figure}
    \centering
    \includegraphics[width=\linewidth]{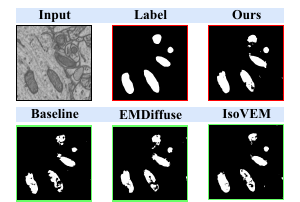}
    \caption{Visual comparison of mitochondria segmentation on the EPFL dataset.}
    \label{fig:seg}
\end{figure}

Qualitative comparisons, shown in Figure~\ref{fig:seg}, corroborate these quantitative findings. The segmentations derived from our method's output exhibit minimal noise and superior structural integrity. Compared to the baseline and other methods, our approach ensures the completeness of the mitochondria, with significantly fewer false negatives and discontinuities.
\subsection{Ablation Study}
To validate the efficacy of our key designs, we performed comprehensive ablation studies on the EPFL dataset for $\times$4 reconstruction. We systematically evaluate the contribution of each core component: the Axial-Lateral Chunking Selective Scan Modul (ALCSSM), the Dynamic Weights Aggregation Module (DWAM), and the Momentum Contrast based degradation learning.

As shown in Table~\ref{table:ablation}, each component contributes to the final performance. Replacing our ALCSSM with a baseline continuous scan results in a 0.07 dB PSNR drop. Substituting our DWAM with simple feature summation causes a 0.061 dB decrease. Furthermore, removing the MoCo-based degradation learning leads to a 0.046 dB decline. Collectively, these results affirm that each proposed component contributes to the final state-of-the-art performance.

\begin{figure}
    \centering
    \includegraphics[width=\linewidth]{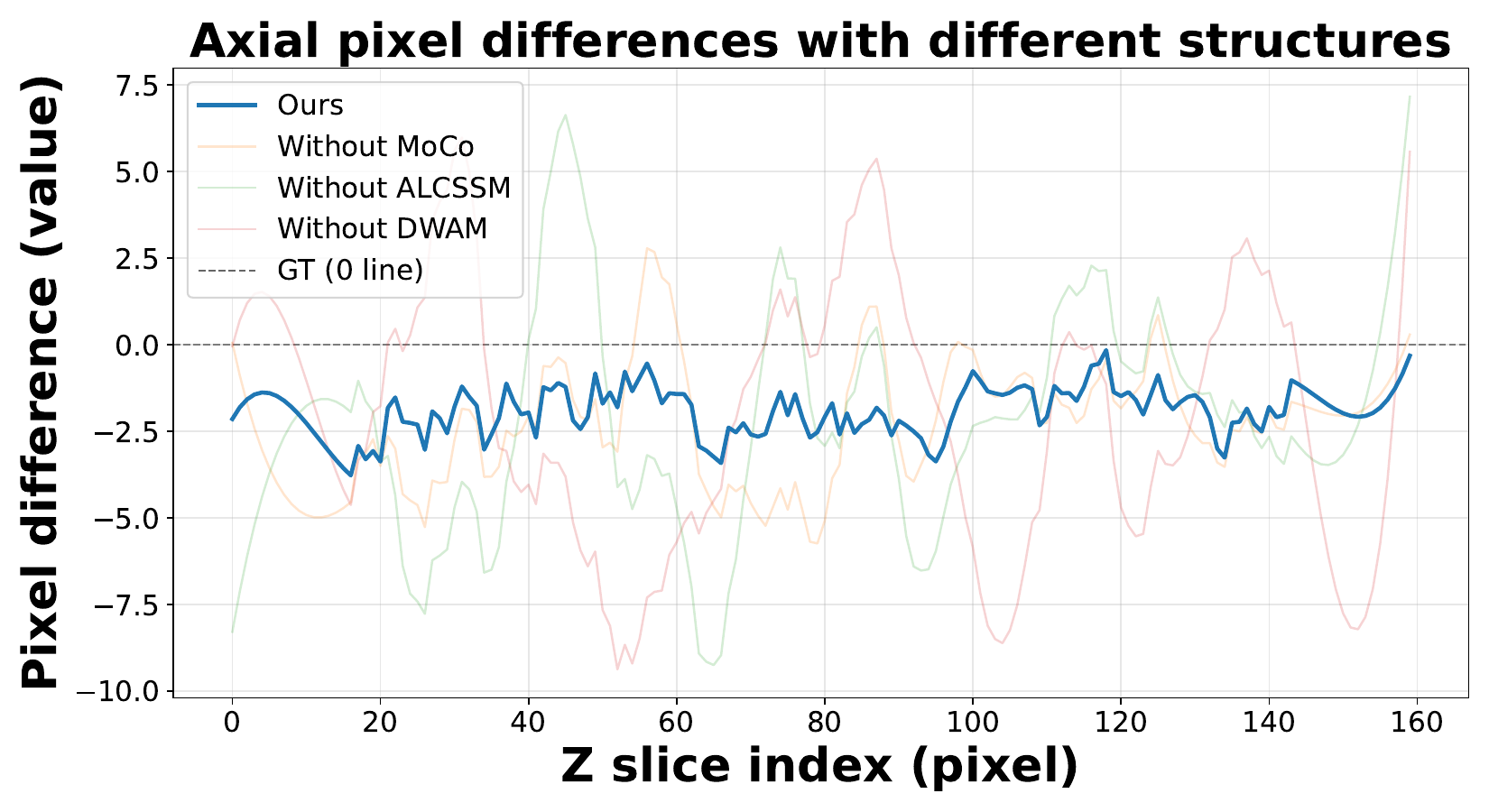}
    \caption{Visualization comparing the distances from GT along the Z-axis for different model architectures on the EPFL dataset.}
    \label{fig:ablation}
\end{figure}

This is further supported by the qualitative analysis in Figure~\ref{fig:ablation}, which plots the axial pixel error. Our full model (``Ours") maintains a minimal and stable deviation from the Ground Truth (0 line). In contrast, models lacking ALCSSM, DWAM or MoCo exhibit significant and erratic pixel differences, with large, fluctuating deviations from the ground truth. This visually confirms that ALCSSM, DWAM and MoCo are essential for establishing the robust axial-lateral consistency that our method promises.


\begin{table}[]
\centering
\caption{Quantitative results of mitochondria segmentation (IoU) on the EPFL dataset. Our method achieves the best performance across all scale factors.}
\label{table:seg}
\begin{tabular}{cccc}
\hline
{\color[HTML]{000000} }                                                                        & \multicolumn{3}{c}{{\color[HTML]{000000} EPFL Dataset}}                                                                  \\
\multirow{-2}{*}{{\color[HTML]{000000} \begin{tabular}[c]{@{}c@{}}Metrics\\ IoU\end{tabular}}} &  $\times4$              &  $\times8 $            & $\times10 $            \\ \hline
{\color[HTML]{000000} Baseline}                                                                & {\color[HTML]{000000} 0.6889}          & {\color[HTML]{000000} 0.6111}          & {\color[HTML]{000000} 0.6099}          \\
{\color[HTML]{000000} IsoVEM}                                                                  & {\color[HTML]{000000} 0.7351}          & {\color[HTML]{000000} 0.6834}          & {\color[HTML]{000000} 0.6802}          \\
{\color[HTML]{000000} EMDiffuse}                                                               & {\color[HTML]{000000} 0.7057}          & {\color[HTML]{000000} 0.6197}          & {\color[HTML]{000000} 0.6151}          \\
{\color[HTML]{000000} Ours}                                                                    & {\color[HTML]{000000} \textbf{0.7464}} & {\color[HTML]{000000} \textbf{0.6975}} & {\color[HTML]{000000} \textbf{0.6927}} \\
{\color[HTML]{000000} Isotropic}                                                               & \multicolumn{3}{c}{{\color[HTML]{000000} 0.7496}}                                                                        \\ \hline
\end{tabular}
\end{table}

\begin{table}[]
\centering

\caption{Ablation study of the proposed ALCSSM, DWAM, and MoCo modules on the EPFL dataset at a scale factor of ×4, evaluated by PSNR and SSIM metrics.}
\label{table:ablation}
\begin{tabular}{ccccc}
\hline
\multirow{2}{*}{ALCSSM} & \multirow{2}{*}{DWAM} & \multirow{2}{*}{MoCo} & \multicolumn{2}{c}{Metrics}   \\ 
                        &                       &                       & PSNR          & SSIM          \\ \hline
\checkmark              &                       & \checkmark            & 29.381         & 0.7695          \\
                        & \checkmark            & \checkmark            & 29.372          & 0.7684          \\
\checkmark              & \checkmark            &                       & 29.396          & 0.7699          \\
\checkmark              & \checkmark            & \checkmark            & \textbf{29.442} & \textbf{0.7707} \\ \hline
\end{tabular}
\end{table}

\section{Conclusion}
In this work, we presented VEMamba, a self-supervised framework that pioneers the Mamba architecture for efficient and effective VEM isotropic reconstruction. Our core innovations are the Axial-Lateral Chunking Selective Scan Module (ALCSSM), which establishes robust 3D axial-lateral consistency, and the Dynamic Weights Aggregation Module (DWAM), which adaptively fuses multi-directional representations. These two modules jointly realize 3D dependency reordering, ensuring globally consistent volumetric reconstruction. We further proposed a realistic degradation simulation strategy, leveraging MoCo to inject degradation-aware knowledge into the network. Extensive experiments demonstrated that VEMamba achieves highly competitive performance, surpassing existing methods in the majority of quantitative metrics, qualitative visual fidelity, and downstream segmentation tasks. Critically, VEMamba accomplishes this with significantly greater computational efficiency, requiring fewer parameters and a lower computational budget than competing models.

\section*{Acknowledgement}
This work was supported by the Natural Science Foundation of China (No. 62272227).

{
    \small
    
    \bibliographystyle{ieeenat_fullname}
    \bibliography{main}
}


\end{document}


\clearpage
\setcounter{page}{1}
\maketitlesupplementary

\section{Introduction to Mamba}
\subsection{State Space Models}
\textbf{Continuous-Time and Discretized SSMs.} SSMs~\cite{ssm1,mamba} are inspired by classical state space systems that model a continuous-time signal $x(t)$ through a latent state $h(t) \in \mathbb{R}^N$ to produce an output $y(t)$. The system dynamics are governed by a linear Ordinary Differential Equation:
\begin{equation}
h'(t) = A h(t) + B x(t), \quad y(t) = C h(t),
\end{equation}
where $A \in \mathbb{R}^{N \times N}$, $B \in \mathbb{R}^{N \times 1}$, and $C \in \mathbb{R}^{1 \times N}$ are the state, input, and output matrices, respectively.

To be applied to discrete data $x_t$, this continuous system must be discretized. This is achieved by introducing a time-step parameter $\Delta$ and transforming $(A, B)$ into discrete parameters $(\overline{A}, \overline{B})$ using a discretization rule, such as the Zero-Order Hold (ZOH):
\begin{equation}
\overline{A} = e^{\Delta A}, \quad 
\overline{B} = (\Delta A)^{-1} (e^{\Delta A} - I) \Delta B,
\label{eq2}
\end{equation}
This results in a discrete-time formulation that can be computed as a linear recurrence:
\begin{equation}
\label{eq3}
h_t = \overline{A} h_{t-1} + \overline{B} x_t, \quad y_t = C h_t.
\end{equation}

\textbf{The Convolutional Duality of LTI SSMs.} A critical property of these Linear Time-Invariant (LTI) models (where $A, B, C, \Delta$ are fixed) is their dual nature. While Eq.~\eqref{eq3} represents the system as a recurrent (RNN) model, it can also be expressed as a discrete convolution $y = x * \overline{K}$. By unrolling the recurrence, the entire output sequence can be computed in parallel by convolving the input $x$ with a global convolutional kernel $\overline{K}$ of length $L$:
\begin{equation}
y = x * \overline{K}, \quad  
\overline{K} = (C\overline{B} , C\overline{AB} , \dots , C\overline{A}^{L-1}\overline{B})
\label{eq4}
\end{equation}
This convolutional representation is highly efficient for parallel training, while the recurrent representation is efficient for auto-regressive inference.

\textbf{Mamba: Selective State Space Models.} The LTI property, while efficient, is also a limitation. The system's dynamics are fixed and independent of the input, constraining its ability to model content-dependent phenomena. Mamba addresses this by introducing a selective mechanism, making the system parameters time-varying and input-dependent.

Specifically, Mamba parameterizes the key system matrices $B$ and $C$, as well as the time-step $\Delta$, as functions of the current input $x_t$:
\begin{equation}
\Delta_t = s_{\Delta}(x_t), \quad B_t = s_B(x_t), \quad C_t = s_C(x_t)
\end{equation}

where $s_{\Delta}, s_B, s_C$ are typically small neural networks. This transforms the LTI recurrence of Eq.~\eqref{eq3} into a time-varying system. The discrete parameters $\overline{A}_t$ and $\overline{B}_t$ are recomputed at each step using the dynamic $\Delta_t$ and $B_t$ (derived from Eq.~\eqref{eq2}):
\begin{equation}
h_t = \overline{A}_t h_{t-1} + \overline{B}_t x_t
,\quad
y_t = C_t h_t
\end{equation}
This simple modification is profoundly impactful. By making the state transitions conditional on the input, the model can selectively choose what information to propagate in its state $h_t$ and what to forget. This input-dependent dynamic breaks the time-invariance necessary for the global convolution (Eq.~\eqref{eq4}), but grants the model significantly enhanced expressive power. This time-varying recurrence is computed efficiently using hardware-aware parallel scan algorithms, retaining the computational benefits of SSMs while enabling content-based reasoning.

\subsection{SSMs for Vision}
The recent emergence of State Space Models (SSMs)~\cite{supply1,supply3}, particularly Mamba~\cite{mamba}, has presented a compelling alternative to Transformers~\cite{supply4} for long-sequence modeling. Mamba's core innovation, the Selective Scan (S6) module, achieves linear-time complexity while effectively capturing long-range dependencies, addressing the quadratic complexity challenge that hinders Transformers in high-resolution tasks. This efficiency has spurred its rapid adoption across diverse computer vision domains, spanning both high-level tasks~\cite{supply7,supply6,supply5} and low-level image processing~\cite{supply10,supply9,supply8}.

However, a fundamental challenge remains in adapting the inherently 1D SSM mechanism to process 2D spatial data. A common paradigm, therefore, involves flattening 2D images into multiple 1D sequences. Pioneering works like Vim~\cite{supply7} and VMamba~\cite{supply6} established this approach. Vim~\cite{supply7} introduced a bidirectional Mamba block, processing image patches sequentially and demonstrating significant speed and memory advantages over Vision Transformers (ViTs) at high resolutions. Concurrently, VMamba~\cite{supply6} proposed a ``Cross-Scan'' strategy, which flattens the input along four directions (row-wise and column-wise, both forward and backward) to integrate spatial context.

Building on these foundational methods, subsequent research has explored a variety of alternative scanning strategies to better align with the 2D structure of images. For instance, continuous scanning paths~\cite{supply11} and local four-directional scans~\cite{supply5} were introduced to enhance structural continuity and local feature acquisition. Other efforts, such as EfficientVMamba~\cite{supply12}, utilizes skip sampling to improve scanning efficiency.

Despite this rapid progress, these methods are fundamentally designed for 2D images and face critical limitations when applied to volumetric data. Existing scanning strategies often disrupt the spatial structure coherence ~\cite{supply10,supply9}, a problem that is exacerbated when naively extended to 3D volumes. While some models like MambaIR~\cite{supply9} and UVM-Net~\cite{supply10} attempt to re-introduce locality using additional CNN layers, this compromises the computational efficiency inherent to the Mamba architecture. Furthermore, these manually designed, fixed scanning paths~\cite{supply5} lack the adaptability to model the complex, anisotropic relationships inherent in VEM data. This often results in a failure to preserve consistency between the high-resolution lateral planes and the sparsely sampled axial dimension. Our work, VEMamba, is designed to overcome these specific challenges by introducing a 3D-native scanning mechanism that explicitly enforces axial-lateral consistency.

\section{Details of VEMamba}
\subsection{Training Strategy}
The training of VEMamba is conducted through a designed two-stage strategy.

\textbf{Stage 1: Degradation Representation Learning.} The initial stage is dedicated to training the degradation encoder. The primary objective of this stage is to learn a robust representation of the degradation characteristics present in the anisotropic data. For this purpose, we adopt the hyperparameter configuration from the CDFormer~\cite{CDFormer}. This pre-training phase enables our model to effectively capture the complex transformations, such as blur and noise, that differentiate the low-resolution inputs from their high-resolution counterparts.

\textbf{Stage 2: Isotropic Reconstruction.} Upon the completion of the encoder training, we freeze its weights to preserve the learned degradation knowledge. Subsequently, the main backbone of the VEMamba model is trained end-to-end. In this stage, the model learns to perform the core task of isotropic reconstruction, guided by the total loss function $\mathcal{L}_{\text{total}}$, as formulated in the main paper. This two-stage approach ensures that the reconstruction process is explicitly conditioned on the learned degradation features, leading to a more targeted and effective restoration.

\subsection{Inference Speed}
We evaluate the inference speed and GPU memory consumption for different methods, with results presented in Table \ref{table:speed}. As shown, our VEMamba demonstrates the highest memory efficiency, consuming only 3308 MiB of GPU memory, which is slightly lower than IsoVEM (3438 MiB) and significantly less than EMDiffuse (6660 MiB).
\begin{table}[h]
\centering
\caption{Inference speed (volumes/s) and GPU memory usage (MiB) comparison on a NVIDIA RTX 4090 with batch size 1.}
\label{table:speed}
\begin{tabular}{ccc}
\hline
Method    & GPU Memory$\downarrow$ & Inference Speed $\uparrow$ \\ \hline
IsoVEM    & 3438                     & 4.524                        \\
EMDiffuse & 6660                     & 0.014                        \\
Ours (VEMamba)     & 3308                     & 2.295                        \\ 
Ours (U-Net)      & 2036                     & 4.598                        \\

\hline
\end{tabular}
\end{table}

Although IsoVEM achieves a higher throughput (4.524 vol/s) than VEMamba (2.295 vol/s), its speed advantage mainly comes from its U-Net–based Transformer architecture, which processes image patches in parallel while progressively downsampling feature maps to reduce computation.

In contrast, VEMamba performs sequential state-space modeling directly on full-resolution features, making it inherently slower. To verify the effect of architectural choices, we also implement a U-Net version of VEMamba, which is an ablation variant and attains higher speed (4.598 vol/s) and lower memory usage (2036 MiB) at the cost of a slight performance drop. Since high reconstruction quality is critical in medical imaging, we prioritize accuracy over speed. Improving the performance of the U-Net variant will be an important direction for future work.

\subsection{Loss Function Selection}
The selection of our composite loss function, $\mathcal{L}_{\text{total}}$, was guided by an empirical study to determine the most effective objective for VEM isotropic reconstruction. Our final formulation is a weighted sum of the $\mathcal{L}_1$ loss and the Structural Similarity Index (SSIM) loss:
\begin{equation}
\mathcal{L}_{\text{total}} = \mathcal{L}_{1}(Y, \hat{Y}) + \mathcal{L}_{\text{SSIM}}(Y, \hat{Y})
\end{equation}
where $Y$ and $\hat{Y}$ represent the ground-truth and the reconstructed volumes, respectively.

To validate this choice, we conducted an ablation study comparing three distinct loss formulations: (1) a pure $\mathcal{L}_1$ loss, (2) a combination of $\mathcal{L}_1$, SSIM, and the LPIPS loss, and (3) our proposed $\mathcal{L}_1 + \mathcal{L}_{\text{SSIM}}$ combination. The quantitative results are presented in Table~\ref{table:lossfunction}. As the results demonstrate, our selected loss function achieves superior performance in terms of both PSNR and SSIM, validating its suitability for preserving both pixel-level accuracy and structural integrity.
\begin{table}[h]
\centering
\caption{Ablation study on the loss function. Our proposed $\mathcal{L}_1 + \mathcal{L}_{\text{SSIM}}$ configuration yields the best results.}
\label{table:lossfunction}
\begin{tabular}{lll}
\hline
\multicolumn{1}{c}{\multirow{2}{*}{Loss   Function}} & \multicolumn{2}{c}{Metrics}       \\
\multicolumn{1}{c}{}                                 & PSNR            & SSIM            \\ \hline
L1                                                   & 29.389          & 0.7659          \\
L1+SSIM+LPIPS                                        & 29.364          & 0.7696          \\
L1+SSIM                                              & \textbf{29.442} & \textbf{0.7707} \\ \hline
\end{tabular}
\end{table}

\section{Details of MoCo}
\subsection{Structure Detail}
In our framework, the degradation representation is extracted using an encoder module within the Momentum Contrast~\cite{moco} setup. The architectural design of this encoder is crucial for effectively capturing the nuanced features of various degradation types. This section provides a detailed breakdown of its structure.

\begin{figure}[h]
    \centering
    \includegraphics[width=\linewidth]{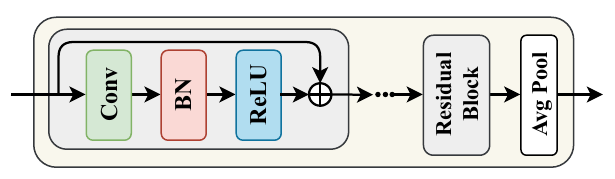}
    \caption{The structure of MoCo encoder. It comprises a convolutional layer, a Batch Normalization layer, and a ReLU activation function, integrated with a skip connection.}
    \label{fig:encoder}
\end{figure}

The encoder is constructed by serially stacking eight identical residual blocks, followed by a final average pooling layer to produce the feature vector. The structure of Encoder is illustrated in Figure~\ref{fig:encoder}.

Specifically, within each block, the input feature map first passes through a convolutional layer (Conv). The output is then normalized using a Batch Normalization (BN) layer to stabilize the training process and accelerate convergence. Following normalization, a Rectified Linear Unit (ReLU) activation function is applied to introduce non-linearity. A characteristic of this architecture is the residual connection, where the input to the block is added element-wise to the output of the ReLU activation. The sequential arrangement of these eight blocks allows the encoder to progressively build a rich and hierarchical representation of the input degradation patterns.

\subsection{Hyperparameter Selection}
To effectively simulate realistic anisotropic data for our self-supervised training paradigm, we employ a degradation pipeline that incorporates Gaussian blur, downsampling, and noise injection. The parameters governing this process are crucial for the model's ability to generalize to real-world data.

Following the methodology established in DiffuseEM~\cite{DiffuseEM}, our primary configuration for the degradation simulation utilizes a Gaussian blur kernel with a filter size of $f=8$ and a standard deviation of $\sigma=4$. To substantiate this choice, we performed a comprehensive ablation study, comparing our selected parameters against several alternatives, including a baseline with only downsampling. The results, summarized in Table~\ref{table:degradation}, clearly indicate that the configuration with $f=8$ and $\sigma=4$ yields the highest PSNR and SSIM scores. This confirms its effectiveness in generating a challenging yet representative training set for our reconstruction task.
\begin{table}[h]
\centering
\caption{Ablation study on degradation simulation parameters. The setting of $f=8, \sigma=4$ provides the best performance.}
\label{table:degradation}
\begin{tabular}{lll}
\hline
\multicolumn{1}{c}{\multirow{2}{*}{Degradation}} & \multicolumn{2}{c}{Metrics}       \\
\multicolumn{1}{c}{}                             & PSNR            & SSIM            \\ \hline
Baseline (downsample)                            & 28.997          & 0.7532          \\
f=8, $\sigma$ = 2                                       & 29.384          & 0.7628          \\
f=8, $\sigma$ = 6                                       & 29.401          & 0.7699          \\
f=10, $\sigma$ = 4                                      & 29.393          & 0.7683          \\
f=6, $\sigma$ = 4                                       & 29.389          & 0.7645          \\
f=8, $\sigma$ = 4                                       & \textbf{29.442} & \textbf{0.7707} \\ \hline
\end{tabular}
\end{table}

\subsection{Effectiveness of MoCo}
\begin{figure}[ht]
    \centering
    \includegraphics[width=1.1\linewidth]{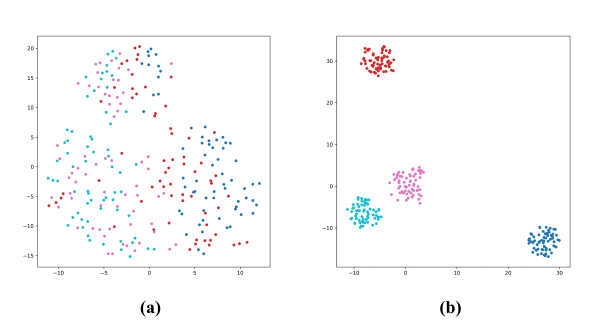}
    \caption{t-SNE visualization of degradation feature distributions. (a) Feature embedding without MoCo training, showing entangled representations where different degradation patterns are indistinguishable. (b) Feature embedding with our MoCo training strategy. The emergence of distinct clusters indicates that the encoder effectively learns to discriminate between different degradation types.}
    \label{fig:tsne}
\end{figure}

To validate the efficacy of our self-supervised degradation learning strategy, we visualize the latent feature distributions of sub-volumes with varying degradation patterns using t-SNE. Figure~\ref{fig:tsne} presents a comparison of the feature space before and after employing Momentum Contrast.

As observed in Figure~\ref{fig:tsne}(a), without contrastive learning, the feature representations of distinct degradation types are heavily entangled and indistinguishable. This lack of separation hinders the network's ability to perceive specific degradation levels. In stark contrast, Figure~\ref{fig:tsne}(b) demonstrates that our MoCo-based training effectively disentangles these representations, resulting in clear, compact clusters corresponding to specific degradation characteristics. This distinct separation confirms that our encoder successfully learns a discriminative degradation prior. Consequently, this enables the Volume Degradation Injection Module (VDIM) to provide precise, degradation-aware modulation to the reconstruction backbone, ensuring robustness against varying anisotropic conditions.

\section{Visualization of EPFL dataset}
\subsection{3D Volume Visualization}
To further complement the 2D slice comparisons presented in the main paper, we provide 3D volumetric visualizations of a representative sub-volume from the EPFL dataset. The following figures demonstrate the performance of our VEMamba against baseline interpolation, IsoVEM, and EMDiffuse across x4, x8, and x10 axial degradation factors.

\begin{figure*}[ht]
    \centering
    \includegraphics[width=1.05\linewidth]{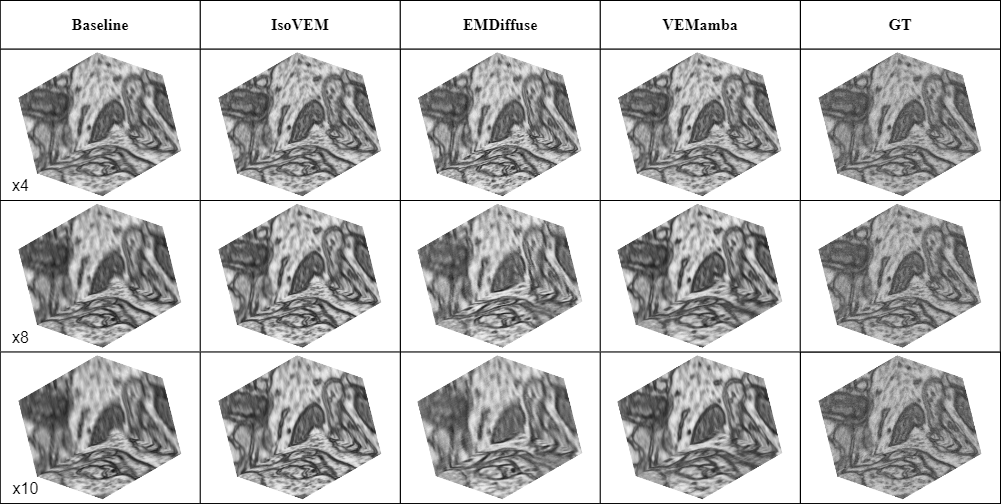}
    \caption{3D volumetric renderings of isotropic reconstruction results on a sub-volume of the EPFL dataset. Our method consistently produces more structurally coherent and detailed reconstructions across all magnification factors (x4, x8, x10) compared to baseline interpolation and competing methods.}
    \label{fig:epfl_3d}
\end{figure*}

Figure \ref{fig:epfl_3d} showcases the 3D renderings of the reconstructed tissue volumes. As can be observed, the baseline interpolation method yields overly smooth and blurry structures, failing to recover fine details, which is consistent with its poor 2D performance. The results from IsoVEM, while improved, exhibit noticeable artifacts such as spurious, disconnected fragments and unnatural surface textures, particularly at higher magnification factors (x8, x10). This aligns with the "hallucinated boundaries" noted in the 2D axial slices (Figure 5 in the main paper). Similarly, EMDiffuse struggles to maintain structural completeness, resulting in volumes that appear eroded or contain discontinuities. In stark contrast, our VEMamba method consistently produces reconstructions with superior structural coherence and surface integrity. The membranes and organelles are rendered with remarkable clarity and continuity, preserving the complex topology of the neural tissue. 

\begin{figure*}[ht]
    \centering
    \includegraphics[width=1.05\linewidth]{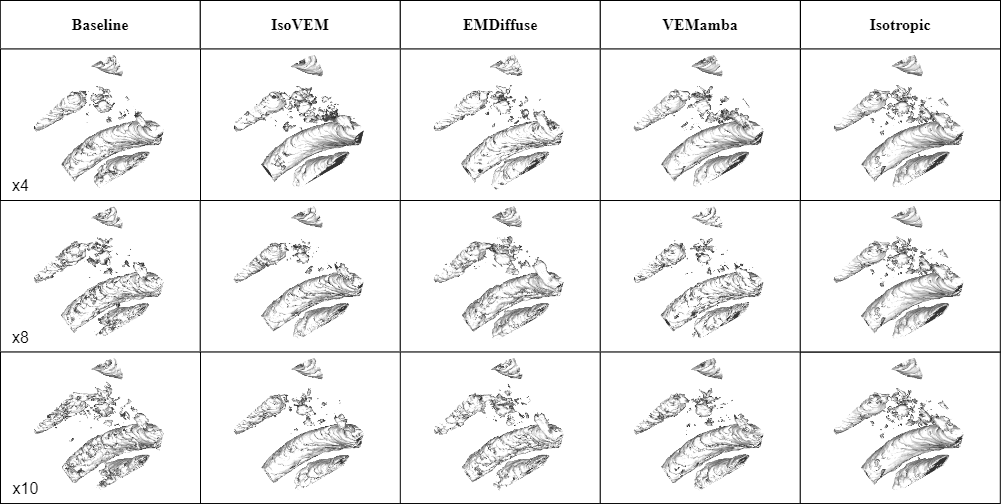}
    \caption{3D visualization of downstream mitochondria segmentation results. The segmentations are based on the reconstructed volumes shown in Figure \ref{fig:epfl_3d}. VEMamba's output enables a significantly more complete and accurate 3D segmentation, with fewer discontinuities and artifacts, underscoring its practical utility for quantitative biological analysis.}
    \label{fig:epfl_seg_3d}
\end{figure*}

Furthermore, the quality of isotropic reconstruction is critical for the accuracy of downstream quantitative analyses. We evaluated this by performing 3D mitochondria segmentation on the reconstructed volumes, as visualized in Figure \ref{fig:epfl_seg_3d}. The segmentations derived from the baseline, IsoVEM, and EMDiffuse reconstructions are heavily fragmented, containing numerous holes and false negatives. These artifacts would severely compromise any subsequent morphological or statistical analysis. The segmentation based on VEMamba's output, however, is significantly more complete and topologically sound. The mitochondria are rendered as continuous, well-defined objects, closely resembling the ground truth. This illustrates that the high fidelity of our reconstruction directly translates into more reliable and accurate results for downstream tasks, corroborating the superior IoU scores reported in the main paper (Table 2).

In summary, these 3D visualizations underscore the superiority of VEMamba in generating high-quality, coherent, and artifact-free isotropic volumes, which is essential for both qualitative inspection and robust quantitative analysis in biomedical research.
\subsection{Axial Section Visualization}
\begin{figure*}[ht]
    \centering
    \includegraphics[width=\linewidth]{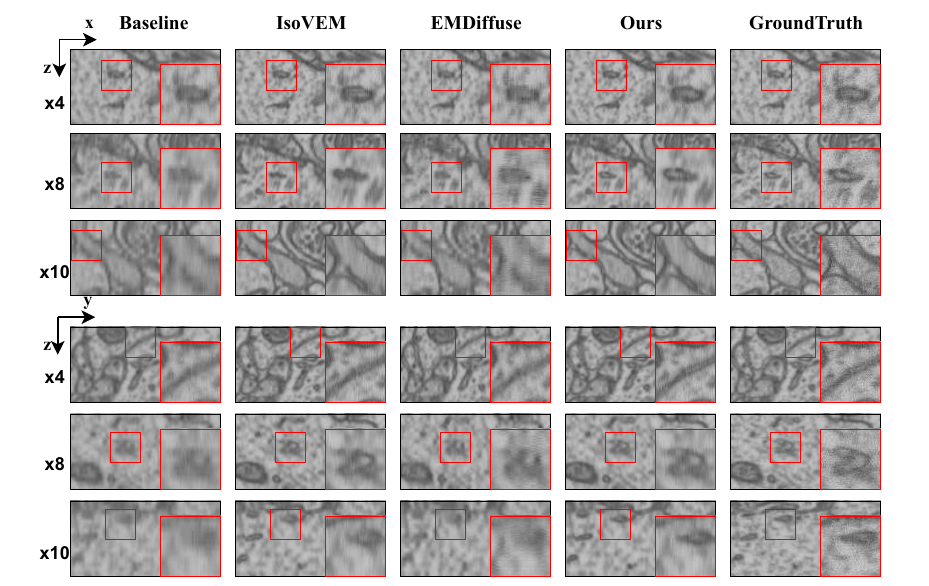}
    \caption{Visual comparison of baseline (Interpolation), IsoVEM, EMDiffuse, VEMamba, and Ground Truth on the EPFL dataset at scale factors of ×4, ×8, and ×10 in the axial (xz and yz) direction.}
    \label{fig:axial}
\end{figure*}

While 2D lateral fidelity is important, the definitive challenge of isotropic reconstruction lies in recovering the missing information along the undersampled axial axis. Figure~\ref{fig:axial} provides a qualitative comparison of reconstructed slices in the axial (xz and yz) planes.

The Baseline method (interpolation) exhibits severe aliasing and blurring, failing to recover any meaningful high-frequency structural details. While IsoVEM recovers sharper textures, it suffers from significant structural inconsistencies; specifically, it generates "hallucinated" boundaries that appear plausible in 2D but result in jagged, discontinuous membrane profiles when viewed axially. EMDiffuse similarly struggles with structural integrity, leading to broken organelle boundaries and noise artifacts.

Conversely, VEMamba demonstrates superior axial consistency. Our method reconstructs smooth, continuous membranes and organelles that closely align with the Ground Truth. The vertical coherence of these structures confirms the effectiveness of the Axial-Lateral Chunking Selective Scan Module (ALCSSM) in effectively modeling 3D spatial dependencies, thereby preventing the slice-to-slice discontinuity observed in competing methods.

\section{Visualization of CREMI dataset}
To complement the quantitative evaluation presented in Table 1 of the main paper, we provide a detailed qualitative comparison of VEMamba's reconstruction performance on the real-world anisotropic CREMI dataset. The following visualizations further substantiate the superior performance of our proposed method.

\begin{figure*}[ht]
    \centering
    \includegraphics[width=\linewidth]{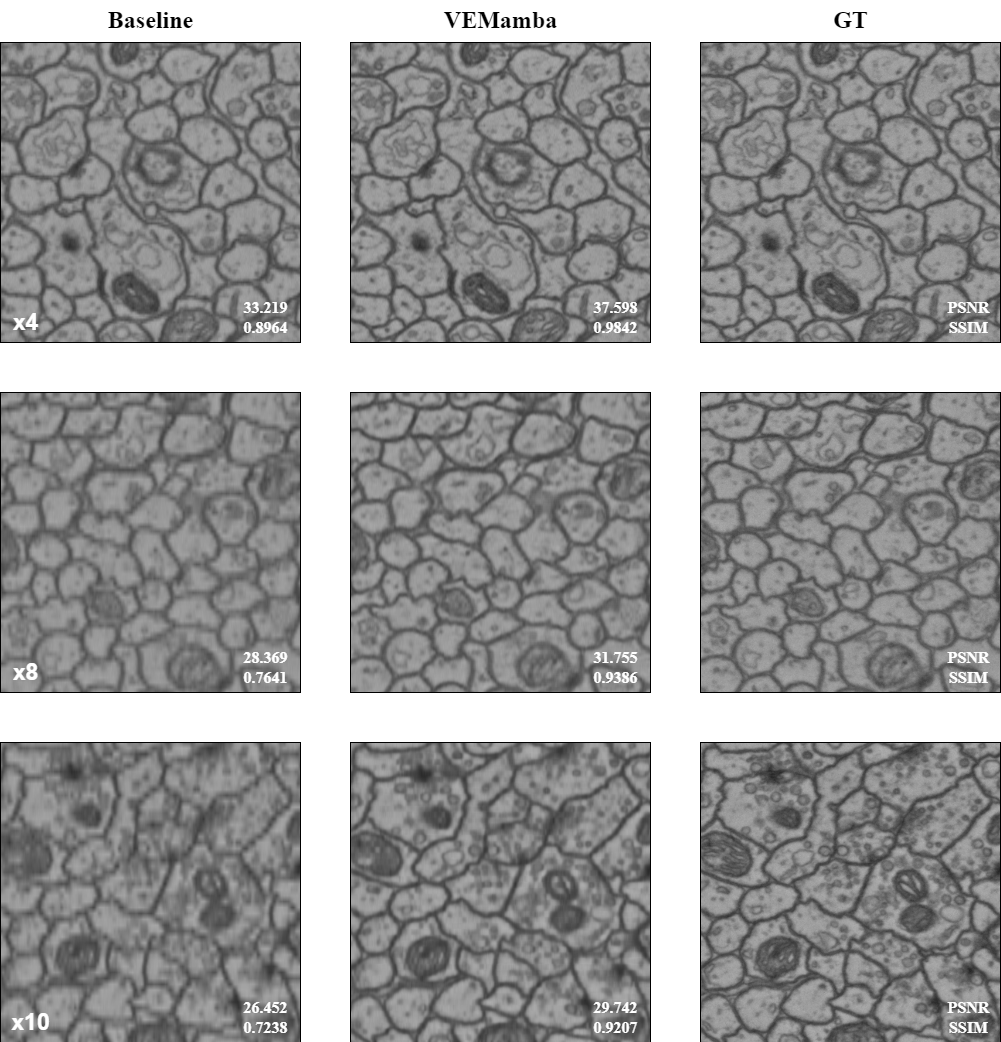}
    \caption{Qualitative comparison on the CREMI dataset (xy-plane). This figure displays 2D lateral sections reconstructed at x4, x8, and x10 degradation factors. VEMamba consistently preserves fine ultrastructural details and sharp boundaries, significantly outperforming the blurry results of the baseline method and closely matching the Ground Truth.}
    \label{fig:cremi_xy}
\end{figure*}

Figure \ref{fig:cremi_xy} illustrates the reconstruction fidelity on the lateral (xy) plane at different degradation factors (x4, x8, and x10). These comparisons are performed by applying our self-supervised framework to the high-resolution lateral sections. It is immediately evident that the Baseline method (representing standard interpolation) fails to recover fine structural details. As the degradation factor increases, the baseline results become progressively blurred, and crucial ultrastructural information. This is particularly noticeable in the x8 and x10 results, where the output is smeared and lacks clarity. In stark contrast, VEMamba demonstrates exceptional performance across all factors. Our model successfully reconstructs sharp, high-frequency details, producing images that are virtually indistinguishable from the Ground Truth (GT). Even at the highly challenging x10 factor, VEMamba robustly restores crisp cell membranes and clearly defined organelles.

\begin{figure*}[ht]
    \centering
    \includegraphics[width=1.05\linewidth]{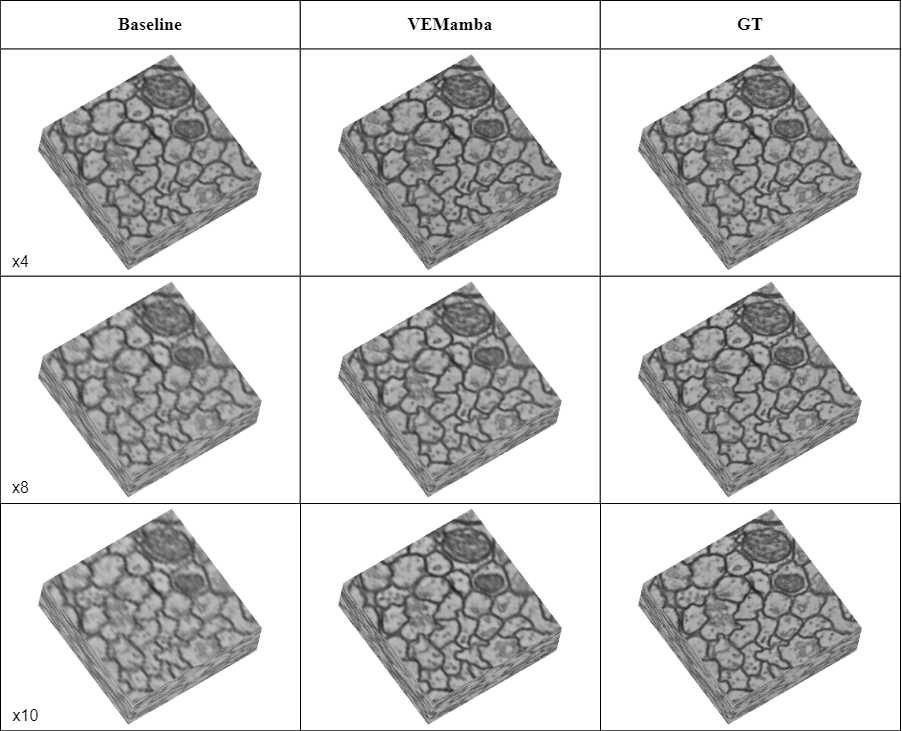}
    \caption{Qualitative comparison of 3D volume rendering on the CREMI dataset. }
    \label{fig:cremi_3d}
\end{figure*}

Beyond 2D fidelity, the ultimate goal of isotropic reconstruction is to achieve 3D structural consistency, which is critical for downstream analyses like segmentation and connectome tracing. Figure \ref{fig:cremi_3d} presents 3D volume.
VEMamba's 3D reconstruction exhibits outstanding spatial consistency. The rendered volume is sharp, and the surfaces of individual cells and organelles are smooth and coherent, faithfully replicating the 3D structure of the Ground Truth. This visualization powerfully demonstrates the success of our Axial-Lateral Chunking Selective Scan Module (ALCSSM) and Dynamic Weights Aggregation Module (DWAM) in capturing long-range, anisotropic spatial dependencies.

In summary, these qualitative results strongly align with our quantitative findings, confirming that VEMamba not only excels at 2D detail restoration but also achieves state-of-the-art 3D consistency, providing a reliable and high-fidelity solution for VEM isotropic reconstruction.


\section{Transferability of VEMamba}
\begin{figure*}[ht]
    \centering
    \includegraphics[width=0.8\linewidth]{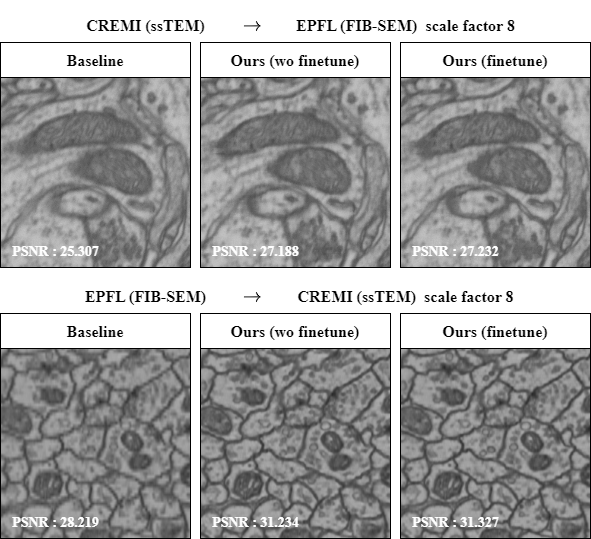}
    \caption{Visual comparison of cross-domain transferability at x8 magnification. Top: CREMI$\rightarrow$EPFL. Bottom: EPFL$\rightarrow$CREMI. VEMamba (wo finetune) achieves results comparable to the fine-tuned model and vastly superior to the baseline, demonstrating strong generalization. }
    \label{fig:transfer}
\end{figure*}

To assess the generalization capability and practical utility of VEMamba, we conduct a challenging transferability study. This is crucial as real-world VEM applications often involve data from diverse sources, exhibiting significant domain shifts due to different imaging modalities, sample preparation protocols, and inherent artifact patterns. We evaluate the model's performance in two scenarios:
\begin{itemize}
    \item \textbf{Zero-shot transfer (Ours (wo finetune))}: The model is trained on one dataset (e.g., EPFL) and directly applied to the other (e.g., CREMI) without any re-training.
    \item \textbf{Fine-tuned transfer (Ours (finetune))}: The model pre-trained on the source dataset is then briefly fine-tuned on a small portion of the target dataset.
\end{itemize}
The quantitative results of this cross-domain evaluation are presented in Table~\ref{table:transfer}, with corresponding visual comparisons provided in Figure~\ref{fig:transfer}.

The results clearly demonstrate VEMamba's robust transferability. As shown in Table~\ref{table:transfer}, our zero-shot model (Ours (wo finetune)) significantly outperforms the baseline across all metrics and transfer directions. For instance, in the EPFL$\rightarrow$CREMI (x8) task, our zero-shot model achieves 30.603 PSNR, substantially surpassing the baseline's 28.129 PSNR. A similar trend is observed in the CREMI$\rightarrow$EPFL (x8) task, where our model (27.392 PSNR) again vastly exceeds the baseline (25.558 PSNR).

Crucially, we observe that fine-tuning the model on the target domain (Ours (finetune)) yields only a \textbf{marginal improvement}. For example, in the EPFL$\rightarrow$CREMI (x8) task, fine-tuning provides only 0.128 dB (30.731 vs. 30.603) of additional gain. This minimal gap strongly indicates that VEMamba, guided by its axial-lateral consistency mechanism, learns fundamental and transferable representations of biological ultrastructures rather than overfitting to modality-specific artifacts.

This observation is visually corroborated in Figure~\ref{fig:transfer}. The reconstructions from our zero-shot model are markedly superior to the baseline, exhibiting significantly enhanced sharpness and structural coherence (e.g., clearer membranes). Furthermore, the visual quality of the ``Ours (wo finetune)" results is virtually indistinguishable from the ``Ours (finetune)" results.

This strong zero-shot performance underscores VEMamba's exceptional generalization. It suggests that our model captures the underlying principles of 3D ultrastructure, making it a robust and practical tool for real-world VEM isotropic reconstruction, even when faced with significant domain shifts.

\begin{table*}[]
\centering
\caption{Quantitative cross-domain transferability results. We compare the baseline, our zero-shot model (wo finetune), and our fine-tuned model (finetune) on tasks transferring between EPFL (FIB-SEM) and CREMI (ssTEM) datasets.}
\label{table:transfer}
\begin{tabular}{cccccccc}
\hline
\multirow{2}{*}{Method}                                                           & \multirow{2}{*}{Metrics} & \multicolumn{3}{c}{EPFL$\rightarrow$   CREMI} & \multicolumn{3}{c}{CREMI$\rightarrow$   EPFL} \\                                                                                  &                          & $\times$4            & $\times$8            & $\times$10           & $\times$4            & $\times$8            & $\times$10           \\ \hline
\multirow{3}{*}{Baseline}                                                         & PSNR                     & 34.219        & 28.129        & 26.554        & 28.407        & 25.558        & 24.583        \\                                                                                  & SSIM                     & 0.9133        & 0.7861        & 0.7379        & 0.7478        & 0.6323        & 0.5891        \\                                                                                  & LPIPS                    & 0.2811        & 0.4445        & 0.5031        & 0.2709        & 0.3955        & 0.4398        \\
\multirow{3}{*}{\begin{tabular}[c]{@{}c@{}}Ours\\      (wo finetune)\end{tabular}} & PSNR                     & 36.804        & 30.603        & 28.932        & 29.345        & 27.392        & 26.438        \\                                                                                  & SSIM                     & 0.9384        & 0.8432        & 0.8018        & 0.7632        & 0.6701        & 0.6355        \\                                                                                  & LPIPS                    & 0.2089        & 0.3363        & 0.3811        & 0.2297        & 0.3239        & 0.3623        \\
\multirow{3}{*}{\begin{tabular}[c]{@{}c@{}}Ours\\      (finetune)\end{tabular}}    & PSNR                     & 36.953        & 30.731        & 29.102        & 29.388        & 27.407        & 26.465        \\                                                                                  & SSIM                     & 0.9403        & 0.8523        & 0.8147        & 0.7676        & 0.6718        & 0.6401        \\                                                                                  & LPIPS                    & 0.2066        & 0.3237        & 0.3696        & 0.2237        & 0.3137        & 0.3542        \\ \hline
\end{tabular}
\end{table*}

{
    \small
    
    \bibliographystyle{ieeenat_fullname}
    \bibliography{main}
}
